\documentclass[12pt]{article}

\usepackage[utf8]{inputenc}
\usepackage[english]{babel}
\usepackage[a4paper,top=2.54cm,bottom=2.54cm,left=1.91cm,right=1.91cm]{geometry}
\usepackage[sfdefault]{carlito}
\usepackage{adjustbox}
\usepackage{amsmath}
\usepackage{soul}
\usepackage{arydshln}
\usepackage{booktabs}
\usepackage{caption}
\usepackage{fancyhdr}
\usepackage{gensymb}
\usepackage{graphicx}
\usepackage{titlesec}
\usepackage[dvipsnames]{xcolor}
\usepackage[colorlinks=true, allcolors=blue]{hyperref}
\usepackage[backend=biber,style=numeric,sorting=none]{biblatex}
\usepackage{amsfonts,amssymb}
\usepackage{algorithmic}
\usepackage{algorithm}
\usepackage{array}
\usepackage{textcomp}
\usepackage{multirow}
\usepackage{subcaption}
\usepackage{float}

\titleformat{\section}[block]
  {\normalfont\Large\bfseries}
  {\thesection.\hspace{10pt}}
  {0pt}
  {}
\titlespacing{\section}{5pt}{1em}{1em}

\titleformat{\subsection}[block]
  {\normalfont\large\itshape}
  {\thesubsection\hspace{10pt}}
  {0pt}
  {}
\titlespacing{\subsection}{5pt}{1em}{1em}

\titleformat{\subsubsection}[block]
  {\normalfont\large\itshape}
  {\thesubsubsection\hspace{10pt}}
  {0pt}
  {}
\titlespacing{\subsubsection}{5pt}{1em}{1em}

\fancypagestyle{firstpagestyle}{
  \fancyhead[C]{\footnotesize\textcolor{gray}{\textit{Computer and Decision Making – An International Journal}
    Volume 2, 2025, 687-707}}
    
    \rfoot{\textcolor{gray}\thepage}
}

\fancypagestyle{otherpagestyle}{
    \lhead{\footnotesize\textcolor{gray}{\textit{Computer and Decision Making – An International Journal}\newline
    Volume 2, (2025) 687-707}}
    \rhead{\empty}
    \rfoot{\textcolor{gray}\thepage}
    \fancyfoot[C]{\empty}
}

\begin{document}
\begin{tabular*}{\textwidth}{ccccr}
\toprule
\multicolumn{2}{p{2in}}{\includegraphics*[width=1.79in, height=1.26in]{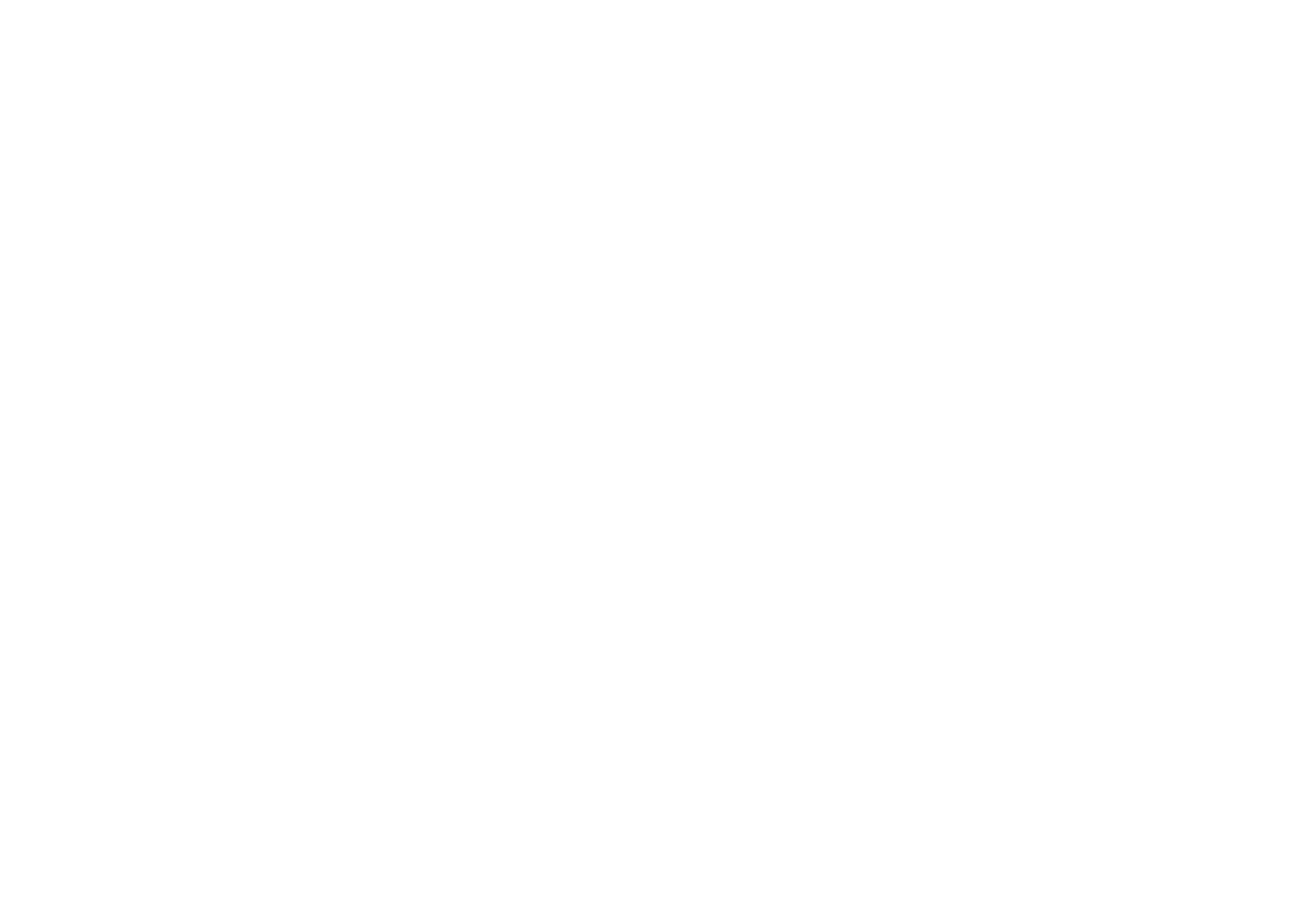}\vspace{-20pt}} & 
\multicolumn{2}{p{3.6in}}{\vspace{-70pt}\hspace{15pt}\textcolor{Blue}{\large \shortstack{Computer and Decision Making\\ An International Journal}}} & \includegraphics*[width=0.8in, height=1.27in]{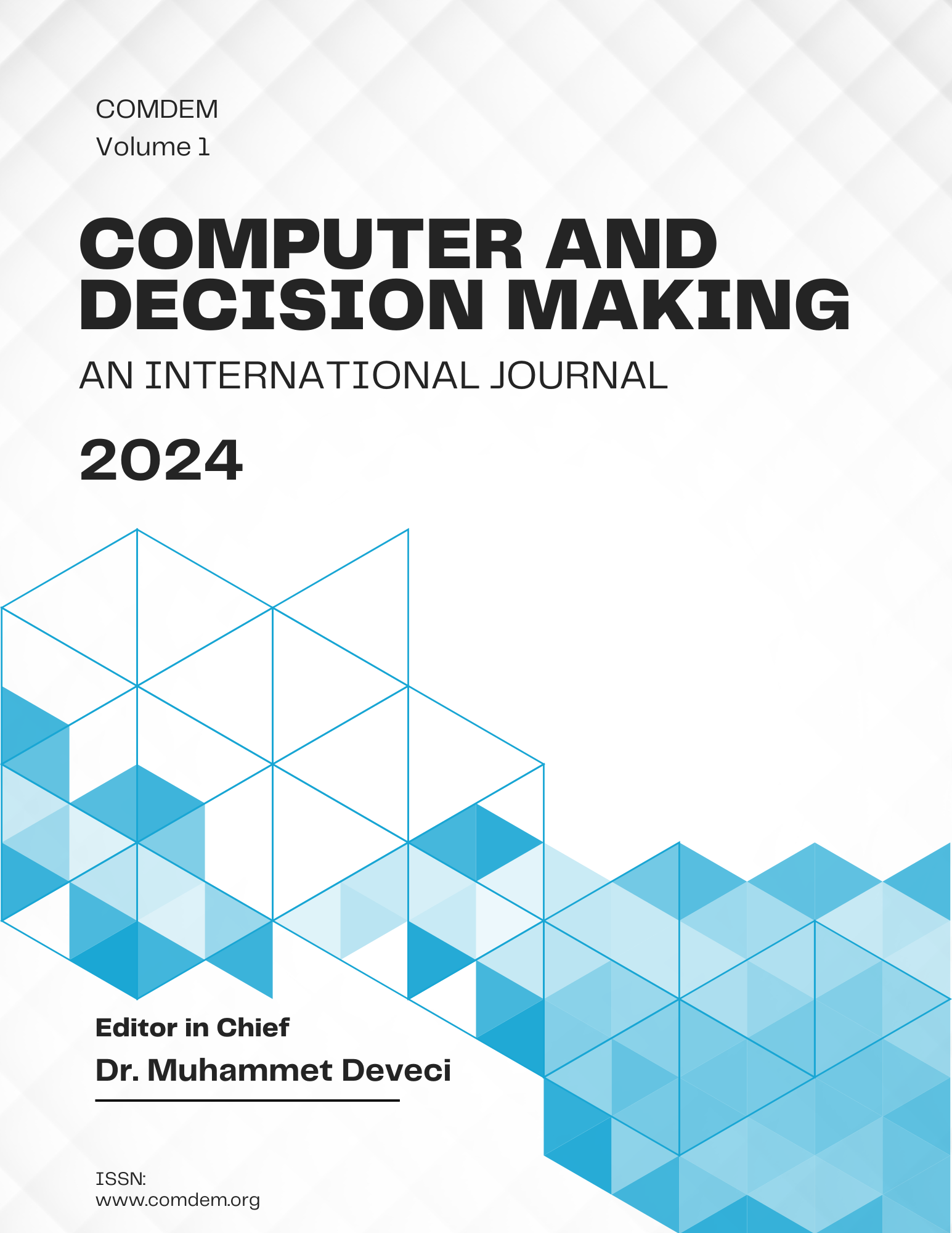}\hspace{20pt}\vspace{-20pt} \\
\multicolumn{2}{p{1in}}{} & \multicolumn{2}{p{2.6in}}{\vspace{-25pt}\footnotesize \textcolor{Blue}{Journal homepage:} \href{https://www.comdem.org}{www.comdem.org} \hfill\break \raggedleft \textcolor{Blue}{eISSN: 3008-1416}\hspace{60pt} \vspace{-20pt}} &  \\
\midrule
\multicolumn{5}{p{1in}}{} \\
\multicolumn{5}{p{\textwidth}}{\vspace{-45pt}\title{\raggedright \textcolor{Blue}{Enhancing Glass Defect Detection with Diffusion Models: Addressing Imbalanced Datasets in Manufacturing Quality Control}}\date{}\maketitle } \\
\multicolumn{5}{p{1in}}{} \\
\multicolumn{5}{p{\textwidth}}{\vspace{-60pt}\author{Sajjad Rezvani Boroujeni${}^{1}$${}^{,}$${}^{2}$${}^{,}$\footnote{}, Hossein Abedi${}^{2}$, Tom Bush${}^{2}$} } \\
 \multicolumn{5}{p{\textwidth}}{\vspace{-30pt} \scriptsize ${}^{1}$ \hspace{15pt}Department of Applied Statistics \& Operations Research (ASOR), Bowling Green State University, Bowling Green, OH, USA} \\
\multicolumn{5}{p{\textwidth}}{\vspace{-35pt} \scriptsize  ${}^{2}$ \hspace{15pt}Data Science Department, Actual Reality Technologies, OH, USA} \vspace{-35pt} \\
 & \multicolumn{4}{p{4.3in}}{} \\ \midrule
\multicolumn{3}{p{2in}}{\textbf{ARTICLE INFO}} & \multicolumn{2}{p{4.5in}}{\textbf{ABSTRACT}} \\ \midrule
\multicolumn{3}{p{2in}}{\textbf{\textit{Article history:\newline }}\scriptsize Received 2 May 2025\newline Received in revised form 1 june 2025\newline Accepted 9 June 2025\newline Available online 10 June 2025} & \multicolumn{2}{p{4.5in}}{\footnotesize Visual defect detection in industrial glass manufacturing remains a critical challenge due to the low frequency of defective products, leading to imbalanced datasets that limit the performance of deep learning models and computer vision systems. This paper presents a novel approach using Denoising Diffusion Probabilistic Models (DDPMs) to generate synthetic defective glass product images for data augmentation, effectively addressing class imbalance issues in manufacturing quality control and automated visual inspection. The methodology significantly enhances image classification performance of standard CNN architectures (ResNet50V2, EfficientNetB0, and MobileNetV2) in detecting anomalies by increasing the minority class representation. Experimental results demonstrate substantial improvements in key machine learning metrics, particularly in recall for defective samples across all tested deep neural network architectures while maintaining perfect precision on the validation set. The most dramatic improvement was observed in ResNet50V2's overall classification accuracy, which increased from 78\% to 93\% when trained with the augmented data. This work provides a scalable, cost-effective approach to enhancing automated defect detection in glass manufacturing that can potentially be extended to other industrial quality assurance systems and industries with similar class imbalance challenges. } \\
\multicolumn{3}{p{2in}}{\vspace{-25pt}\textbf{\textit{Keywords:}}} & \multicolumn{2}{p{2.5in}}{\textbf{}} \\
\multicolumn{3}{p{2in}}{\vspace{-25pt} \footnotesize Defect detection; diffusion models;
data augmentation; glass manufacturing; imbalanced datasets; CNN;
quality control; industrial quality
assurance; generative models;
anomaly detection; machine
learning } & \multicolumn{2}{p{2.5in}}{\textbf{\vspace{-10pt}}} \\ \bottomrule
\end{tabular*}

\footnotetext{\textit{Corresponding author.\newline
E-mail address: sajjadr@bgsu.edu} \\ \newline
https://doi.org/10.59543/comdem.v2i.14391}

\thispagestyle{firstpagestyle}

\section{Introduction}

\hspace{0.63cm}The global manufacturing sector is undergoing a rapid transformation as industries adopt advanced technologies to meet increasing demands for quality, efficiency, and sustainability. In glass container manufacturing—a market valued at USD 106.4 billion in 2021 and projected to reach USD 167.5 billion by 2030—defects such as cracks, bubbles, dimensional deviations, surface imperfections, and contamination pose significant challenges to maintaining product integrity [1]. These defects not only compromise safety but also lead to substantial financial losses due to recalls, material waste, and production downtime. Automated inspection systems, particularly those leveraging deep learning, have demonstrated significant cost savings by reducing inspection errors and improving throughput as shown by Hütten \textit{et al.} [2].

Traditional defect detection methods rely heavily on manual inspections or rule-based machine vision systems, both of which are prone to errors and inefficiencies. Recent advances in artificial intelligence (AI), particularly deep learning models such as convolutional neural networks (CNNs), have demonstrated significant potential in automating defect detection processes [3]. Models like ResNet50V2, EfficientNetB0, and MobileNetV2 achieve high accuracy in identifying visible defects such as cracks and bubbles. However, the performance of AI models is often constrained by data scarcity and class imbalance; rare defects are underrepresented in training datasets due to their infrequent occurrence in real-world production lines. Furthermore, these systems face limitations in detecting subtle or rare defects such as internal contamination caused by silica residues or foreign particulates in recycled glass production.

In the exploration of solutions to these challenges, synthetic data augmentation techniques emerge as a promising approach. Denoising Diffusion Probabilistic Models (DDPMs) represent a significant advancement in this domain due to their capability to generate high-resolution synthetic images with superior detail preservation. Although DDPMs have been successfully applied in fields like medical imaging for anomaly detection by Wolleb \textit{et al.} [4] and addressing imbalanced medical datasets by Khazrak \textit{et al.} [5], their application to industrial glass defect detection remains largely unexplored in the literature.

This research aims to bridge this gap by leveraging DDPMs to enhance defect detection in glass container manufacturing through synthetic data augmentation. The work is structured into two main phases: first, baseline evaluations using popular CNN architectures (ResNet50V2, EfficientNetB0, MobileNetV2) trained on a real-world dataset to establish their performance in detecting defective items; second, DDPM-generated synthetic images are integrated into the training pipeline to address data scarcity and improve dataset balance, enhancing model robustness against defects. The approach is evaluated using standard metrics such as accuracy, precision, recall, F1 scores, and ROC AUC.

The experimental results demonstrate that the integration of synthetic defect images significantly improves model performance across all tested architectures. Particularly notable is the improvement in recall rates—the ability to correctly identify true defects—which increased from 35\% to 65\% for EfficientNetB0, 60\% to 67\% for MobileNetV2, and 80\% to 84\% for ResNet50V2, all while maintaining perfect precision. This indicates that the approach effectively addresses the class imbalance problem without introducing false positives, a critical requirement in manufacturing quality control.

In summary, this paper makes three key contributions: (i) Novel application of DDPMs: Demonstration of diffusion models for generating high-fidelity glass defect images that effectively supplement imbalanced manufacturing datasets. (ii) Comprehensive model evaluation: Systematic analysis of how synthetic data augmentation affects the performance of three distinct CNN architectures in glass defect detection. (iii) Improved detection of minority class: Significant increases in recall and F1 scores for defect detection while maintaining zero false positives, addressing a critical challenge in manufacturing quality control.

By addressing the challenges of class imbalance and data scarcity, this research provides a template for AI-driven quality assurance in glass manufacturing, a critical step toward sustainable, efficient production systems. In the last paragraph of the introduction section, we highlight the gap in applying DDPMs to industrial glass defect detection, the significance of addressing class imbalance in manufacturing quality control, and the objective of developing a synthetic data augmentation approach that improves defect detection performance across multiple CNN architectures.

\thispagestyle{otherpagestyle}
\section{Related Work}

\subsection{Traditional and Modern Defect Detection in Glass Manufacturing}

\hspace{0.63cm}The journey of quality control in glass manufacturing has evolved tremendously over time. Human inspectors—once the backbone of quality control—have increasingly given way to AI-driven systems. While human experts could visually identify defects, they inevitably faced issues with consistency, fatigue, and subjective judgments [6]. The first wave of automated systems emerged in the 1980s-1990s with basic image processing techniques like edge detection, thresholding, and template matching. Yet these conventional approaches often faltered when confronted with varying lighting conditions, complex reflections on glass surfaces, and the diverse ways defects can manifest as described by Malamas \textit{et al.} [7].

Glass products in manufacturing present unique inspection challenges due to their geometric properties. Their curved or cylindrical surfaces create complex reflection patterns, and the semi-transparent nature of glass makes defect detection particularly challenging. Traditional inspection systems required multiple cameras at different angles and sophisticated image stitching algorithms. Despite these measures, internal defects like bubbles, stones (unmelted raw materials), and cord (glass inhomogeneity) remained elusive to conventional vision systems [6].

Recent years have witnessed a significant transformation with CNN applications in the field. For glass manufacturing specifically, Zhou \textit{et al.} [8] developed a visual attention-based framework that successfully detects subtle defects on glass surfaces by combining wavelet transform with deep learning models. Their approach effectively identified complex defects that conventional systems often missed, while reducing computational complexity. Yet these deep learning approaches still face significant implementation hurdles in production environments—primarily due to class imbalance, as defective products typically make up only 1-5\% of production output [2].

\subsection{Addressing Class Imbalance and Data Augmentation}

\hspace{0.63cm}The manufacturing community has explored several strategies to address class imbalance in industrial quality control: Resampling techniques, where Buda \textit{et al.} [9] found that oversampling defective examples improved detection accuracy for rare defects by about 15\%. Cost-sensitive learning approaches, where Khan \textit{et al.} [10] demonstrated that weighted loss functions boosted recall for defect detection in transparent materials by up to 23\% without significantly increasing false positives. Ensemble methods, which combine multiple models trained on different data subsets and have proven effective for handling diverse defect types [11]. Data augmentation, which creates synthetic data to increase dataset diversity without collecting more real-world samples [12].

Traditional data augmentation techniques like geometric transformations and color adjustments have shown limited effectiveness for glass manufacturing defects. These methods simply cannot capture the unique optical properties of glass materials. Early approaches using Generative Adversarial Networks (GANs) for synthetic defect generation faced significant challenges. GANs often encountered mode collapse, where the generator produces limited varieties of samples, failing to capture the full diversity of defect types. Additionally, training instability issues were particularly pronounced when dealing with transparent materials like glass, where subtle variations in refraction patterns and transparency need to be accurately reproduced. These GANs frequently failed to generate realistic optical characteristics such as proper light transmission through cracks or accurate representation of contamination within the glass matrix.

Diffusion models have recently emerged as a promising alternative. For glass defect image generation, DDPMs offer several compelling advantages: they excel at high-fidelity generation that captures subtle variations in transparency and texture; they offer better training stability compared to GANs [13]; and they allow for conditioning generation on specific defect types [14-16]. In related research areas, Khazrak \textit{et al.} [5] demonstrated the effectiveness of DDPMs for addressing imbalanced medical imaging datasets—findings that translate well to glass defect detection since both domains involve identifying subtle anomalies in visually complex contexts.

\subsection{Transfer Learning for Glass Defect Detection}

\hspace{0.63cm}Transfer learning has proven particularly valuable for glass defect detection, especially when annotated data is limited [17]. In glass manufacturing specifically, pre-trained architectures like ResNet50V2, EfficientNetB0, and MobileNetV2 have been successfully adapted to detect surface defects, with each offering different trade-offs between accuracy and computational requirements [18-20].

Ferguson \textit{et al.} [21] showed that transfer learning could reduce required training data by up to 70\% compared to training from scratch. However, they noted an important limitation: transfer learning effectiveness is often constrained by dataset imbalances. This observation highlights the potential benefit of combining transfer learning with synthetic data augmentation techniques like diffusion models to tackle the persistent challenges in glass quality control systems.

\section{Methodology}

\hspace{0.63cm}This section details the approach to addressing class imbalance in glass defect detection using diffusion models. As illustrated in Figure 1, the methodology encompasses data preparation, DDPM training, synthetic data generation, CNN model training, and comprehensive evaluation.

\begin{figure}[H]
\centering
\includegraphics[width=\linewidth]{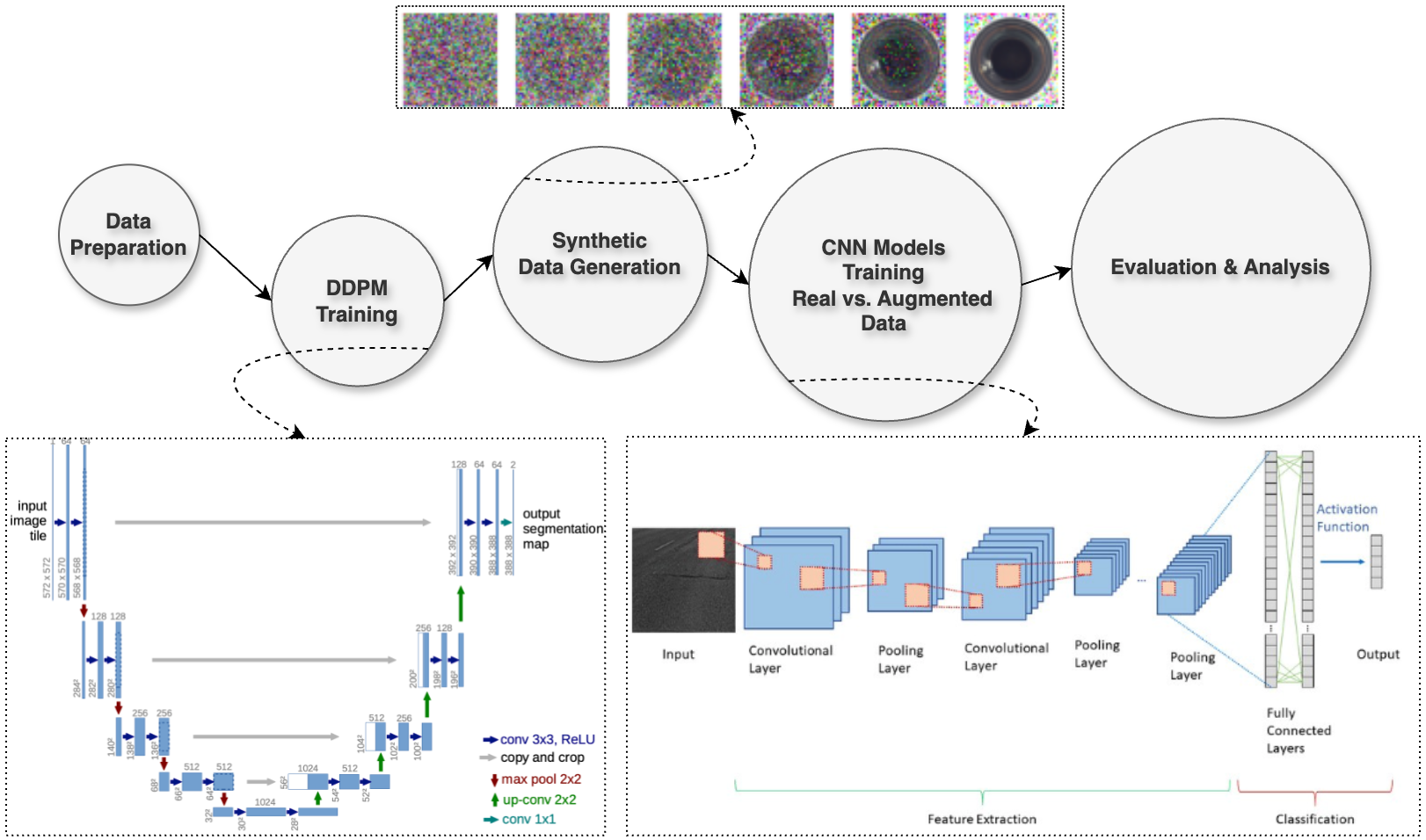}
\caption*{\textbf{Fig. 1.} Overview of the proposed methodology for enhancing glass defect detection. The pipeline consists of five main stages: (1) data preparation using the MVTec glass bottle dataset, (2) training a DDPM on defective samples [22], (3) generating synthetic defective samples to augment the dataset, (4) training CNN models on both original and augmented datasets, and (5) evaluation and feature space analysis}
\end{figure}

\subsection{Dataset Description}

\hspace{0.63cm}The dataset used in this study is from the MVTec Anomaly Detection Dataset [23], which is licensed under the Creative Commons Attribution-NonCommercial-ShareAlike 4.0 International License (CC BY-NC-SA 4.0). This license allows for sharing and adaptation for non-commercial purposes, provided appropriate credit is given and derived works are shared under the same license terms.

The research utilizes a real-world dataset of glass bottle images collected from an operational production line. The dataset consists of 272 images categorized into two classes: Non-Defective: 209 images (76.8\% of the dataset) and Defective: 63 images (23.2\% of the dataset).

This distribution reflects the real-world scenario in manufacturing environments where defective products constitute the minority class. The significant class imbalance (approximately 3:1 ratio) presents a challenge for machine learning models, which tend to favor the majority class during training.

\subsection{Diffusion Model for Synthetic Data Generation}

\hspace{0.63cm}To address the class imbalance, a Denoising Diffusion Probabilistic Model (DDPM) was employed to generate synthetic images of the minority class (defective products). DDPMs operate by learning a reverse diffusion process that gradually transforms noise into structured data [14].

Diffusion models consist of a forward process that gradually adds noise to an image $x_0$ according to a variance schedule $\beta_1,...,\beta_T$, and a learned reverse process that denoises the image. The forward diffusion process gradually adds noise to the data. Mathematically, the forward process at timestep $t$ is defined as:

\begin{equation}
q(x_t|x_{t-1}) = \mathcal{N}(x_t; \sqrt{1-\beta_t}x_{t-1}, \beta_t\mathbf{I})
\end{equation}

The reverse denoising process, which is learned during training, predicts the parameters
\(\mu_\theta(x_t, t)\) and \(\Sigma_\theta(x_t, t)\)
of the Gaussian kernel via a neural network. It can be written as:
\begin{equation}
p_\theta(x_{t-1}|x_t) = \mathcal{N}(x_{t-1}; \mu_\theta(x_t, t), \Sigma_\theta(x_t, t))
\end{equation}

Advanced implementations of diffusion models have explored numerous refinements to this basic formulation. Karras \textit{et al.} [24] conducted a comprehensive analysis of the design space of diffusion models, proposing improvements to both training and sampling procedures that enhance efficiency and image quality. Their work demonstrates that careful selection of noise schedules and sampling strategies can significantly reduce computational costs while maintaining high-fidelity generation.

Figure 2 illustrates this two-phase process. The forward process (bottom) gradually adds noise to an image until it becomes pure noise, while the backward process (top) learns to reverse this degradation by removing noise step by step to recover the original image.

\begin{figure}[H]
\centering
\includegraphics[width=0.9\linewidth]{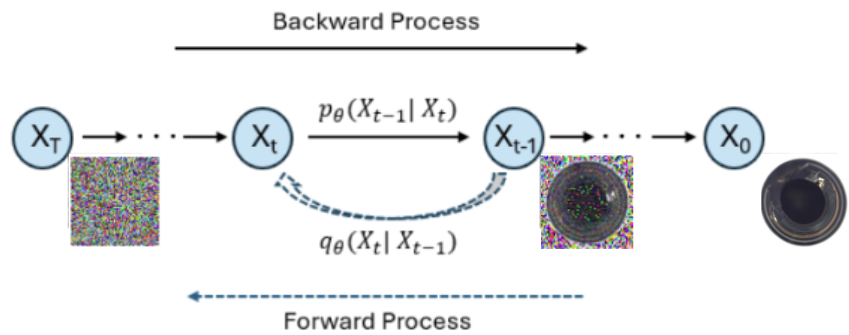}
\caption*{\textbf{Fig. 2.} Illustration of the diffusion model's forward and backward processes. The forward process (bottom, dashed arrow) gradually adds noise to a clean glass bottle image ($X_0$) until it becomes pure noise ($X_T$). The model learns the backward process (top, solid arrow) which denoises the image step by step, converting noise back into a realistic glass bottle image}
\end{figure}

\subsubsection{Dataset Preparation for DDPM Training}

A custom PyTorch dataset class (SingleClassDataset) was created specifically focused on the defective class. This dataset: Loaded images exclusively from the "Defective" folder containing 63 images; Included all PNG and JPG format images; Applied preprocessing transformations including resizing to 128×128 pixels, conversion to tensors, and normalization (mean and standard deviation of 0.5 for each channel); Was configured with a DataLoader using batch size of 8, shuffling enabled, and two worker processes.

No additional data augmentation was applied during DDPM training, allowing the model to learn directly from the raw defective samples without synthetic variations.

\subsubsection{DDPM Architecture and Training}

A UNet-based diffusion model was implemented with the following specifications: Base architecture: UNet2DModel with sample size of 128×128 pixels, input/output channels of 3 (RGB images), multiple layers per block with tailored block output channels, and combination of standard down/up sampling blocks and attention-based blocks. Noise scheduling: DDPMScheduler configured with 14,000 training timesteps and linear beta schedule (ranging from 0.0001 to 0.02). Pipeline: DDPMPipeline combining the UNet model with the noise scheduler. Computation: Model trained on CUDA-enabled GPU.

While our implementation utilized a standard UNet architecture, recent advancements in diffusion model efficiency have introduced alternative approaches. Rombach \textit{et al.} [25] demonstrated that diffusion models operating in the compressed latent space of autoencoders (Latent Diffusion Models) can achieve comparable image quality with significantly reduced computational requirements. This approach is particularly promising for industrial applications where computational resources may be limited.

The training procedure consisted of: Optimizer: AdamW with learning rate of $1\times10^{-4}$. Training duration: 1,300 epochs. Training loop for each batch: Random timestep sampling from the scheduler's range, noise addition to images according to sampled timesteps, noise prediction by the UNet model, MSE loss computation between predicted and actual noise, and backpropagation and optimizer step for weight updates. Monitoring: Loss logging every 50 steps to track convergence. Checkpointing: Saving scheduler configuration and model state dictionary for later inference.

\subsubsection{Synthetic Data Generation Process}

After training, the following process was employed to generate new synthetic defective bottle glass images: Image sampling through sampling an initial noise tensor as the starting point, iteratively refining this tensor using the scheduler's timesteps and the trained UNet model, and post-processing the final images (clamping and scaling) to ensure appropriate pixel value ranges. Output: Saving generated images to a specified output directory. Validation: Visual confirmation of generation quality.

Using this pipeline, diverse synthetic defective glass bottle images that captured various defect types were successfully generated. Figure 3 shows representative examples from this approach.

\begin{figure}[H]
\centering
\includegraphics[width=0.8\linewidth]{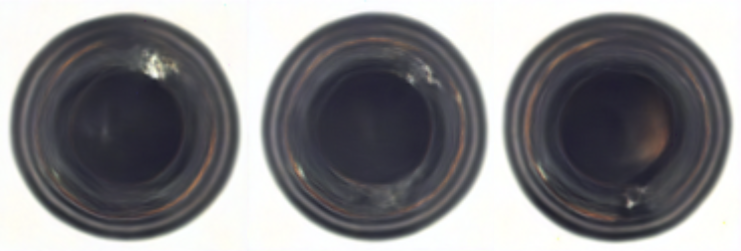}
\caption*{\textbf{Fig. 3.} DDPM-generated synthetic defective bottle glass images captured from a top-down view. The two left images depict breaks and cracks, easily identifiable by jagged and irregular edge contours, while the rightmost image highlights a contamination defect, visible as a brown discoloration concentrated at the bottom of the bottle, an important cue for identifying internal residue or inclusion-related faults}
\end{figure}

In total, 60 new synthetic defective glass images were generated, resulting in an augmented dataset with the following distribution: Non-Defective: 209 images (unchanged) and Defective: 123 images (63 real + 60 synthetic).

This increased the defective class representation from 23.2\% to 37.0\% of the total dataset, reducing the imbalance ratio from approximately 3:1 to 1.7:1. Figure 4 shows a comparison between real non-defective, real defective, and synthetic defective glass bottle images.

\begin{figure}[H]
\centering
\includegraphics[width=0.95\linewidth]{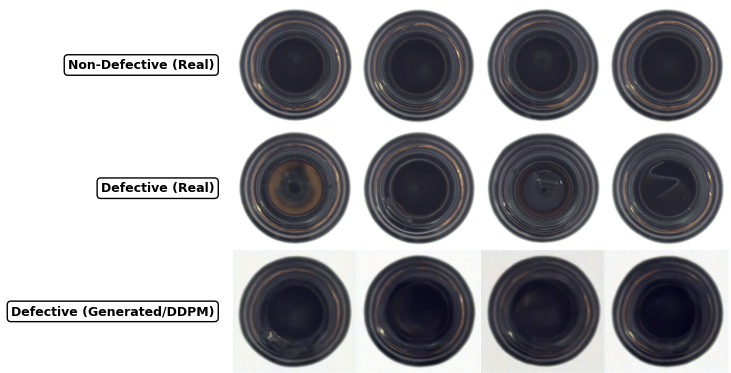}
\caption*{\textbf{Fig. 4.} Comparison of bottle glass samples: (Top row) Real non-defective glass bottles showing consistent clarity and structure. (Middle row) Real defective glass bottles exhibiting various imperfections including discoloration, structural irregularities, breaks, and bottom-of-the-bottle contamination. (Bottom row) DDPM-generated synthetic defective glass bottles that capture similar defect patterns to the real samples, showing clear examples of cracks and contamination localized at the bottle base}
\end{figure}

\subsection{CNN Models for Defect Classification}

\hspace{0.63cm}Three popular CNN architectures were evaluated, each with different characteristics in terms of depth, parameter count, and computational efficiency:

\subsubsection{ResNet50V2}
ResNet50V2 is a deep residual network featuring skip connections that help address the vanishing gradient problem in deep networks [26]. With approximately 25.6 million parameters, it represents a deep and powerful architecture.

\subsubsection{EfficientNetB0}
EfficientNetB0 utilizes compound scaling to balance network depth, width, and resolution, achieving strong performance with fewer parameters (approximately 5.3 million) [19].

\subsubsection{MobileNetV2}
MobileNetV2 employs depthwise separable convolutions and inverted residuals to create a lightweight architecture (approximately 3.5 million parameters) suitable for resource-constrained environments [20].

\subsection{Experimental Setup}

\hspace{0.63cm}For each CNN architecture, two experiments were conducted: (1) Baseline (RealData): Training on the original dataset with 209 non-defective and 63 defective images. (2) Augmented (AugmentedData): Training on the expanded dataset with 209 non-defective and 123 defective images (including DDPM-generated synthetic images).

\subsubsection{Data Preprocessing and Augmentation}

All images were resized to 128×128 pixels and preprocessed according to the requirements of each specific CNN architecture. Standard data augmentation techniques were applied during training to improve model generalization: Random horizontal and vertical flips, Random rotation (±20\%), Random zoom (±20\%), and Random contrast adjustment (±20\%).

These augmentations were applied to both experimental setups, with the key difference being the inclusion of the 60 DDPM-generated images in the AugmentedData experiment.

\subsubsection{Model Training}

For all experiments, transfer learning was employed by using pre-trained ImageNet weights for the backbone networks while training only the classification head. This approach leverages general feature extraction capabilities developed on a large dataset while adapting specifically to the glass defect detection task.

The models were trained with the following specifications: Optimizer: Adam with learning rate 1e-4. Loss function: Binary cross-entropy. Batch size: 32. Maximum epochs: 5. Early stopping: Patience of 5 epochs, monitoring validation loss. Learning rate schedule: ReduceLROnPlateau with factor 0.2 and patience 3.

To further address class imbalance during training, class weights were applied. These weights are calculated as shown in Eq. (3):

\begin{equation}
w_0 = \frac{N_{total}}{N_{NonDefective} \times 1.2}, \quad w_1 = \frac{N_{total}}{N_{Defective} \times 2.1}
\end{equation}

where $N_{total}$ is the total number of samples, $N_{NonDefective}$ is the number of non-defective samples, and $N_{Defective}$ is the number of defective samples. This weighted approach helps the model pay more attention to the minority class during training.

\subsection{Evaluation Metrics}

\hspace{0.63cm}Model performance was evaluated using standard metrics for binary classification, with particular attention to metrics that reflect the ability to identify defective products (the minority class):

Accuracy: Overall proportion of correct predictions
\begin{equation}
\text{Accuracy} = \frac{TP + TN}{TP + TN + FP + FN}
\end{equation}

Precision: Proportion of predicted defects that are actual defects
\begin{equation}
\text{Precision} = \frac{TP}{TP + FP}
\end{equation}

Recall: Proportion of actual defects that are correctly identified (also known as sensitivity)
\begin{equation}
\text{Recall} = \frac{TP}{TP + FN}
\end{equation}

F1 Score: Harmonic mean of precision and recall
\begin{equation}
\text{F1} = 2 \times \frac{\text{Precision} \times \text{Recall}}{\text{Precision} + \text{Recall}}
\end{equation}

ROC AUC: Area under the Receiver Operating Characteristic curve, which plots true positive rate against false positive rate at various threshold settings
\begin{equation}
\text{ROC AUC} = \int_{0}^{1} TPR(FPR^{-1}(t)) dt
\end{equation}
where $TPR$ is the true positive rate (recall) and $FPR$ is the false positive rate given by:
\begin{equation}
\text{FPR} = \frac{FP}{FP + TN}
\end{equation}

In these formulas, $TP$ represents true positives (correctly identified defects), $TN$ represents true negatives (correctly identified non-defective samples), $FP$ represents false positives (non-defective samples incorrectly classified as defective), and $FN$ represents false negatives (defective samples incorrectly classified as non-defective).

Given the industrial context where missing defects (false negatives) can have serious consequences but where excessive false alarms (false positives) can also be costly, the classification threshold was set at 0.4 to balance precision and recall. This approach to threshold optimization for imbalanced datasets aligns with techniques demonstrated in other domains where CNN architectures have been employed for binary classification of complex signals [27].

\section{Results}

\subsection{ResNet50V2 Results}

\begin{table}[H]
\centering
\caption*{\textbf{Table 1}\\ResNet50V2 Performance Comparison}
\vspace{-10pt}
\begin{tabular}{lcc}
\toprule
\textbf{Metric} & \textbf{RealData} & \textbf{AugmentedData} \\
\midrule
Accuracy & 0.78 & 0.93 \\
Precision & 0.53 & 1.00 \\
Recall & 0.80 & 0.84 \\
F1 Score & 0.64 & 0.91 \\
ROC AUC & 0.899 & 0.973 \\
\bottomrule
\end{tabular}
\end{table}

\begin{figure}[H]
\centering
\includegraphics[width=0.9\linewidth]{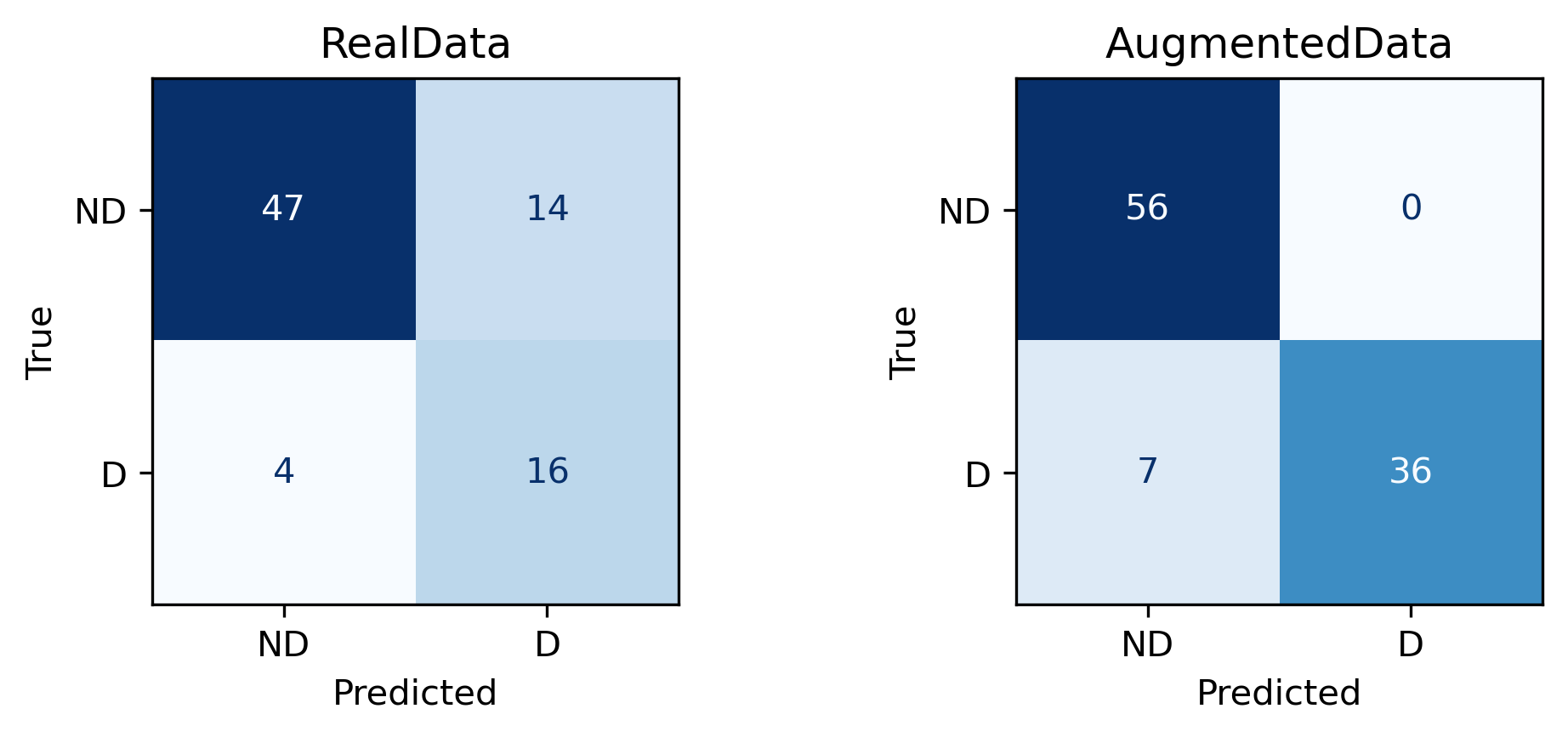}
\caption*{\textbf{Fig. 5.} ResNet50V2 Confusion Matrices for RealData (left) and AugmentedData (right). ND = Non-Defective, D = Defective}
\end{figure}

\hspace{0.63cm}The ResNet50V2 model showed remarkable improvement when trained with the augmented dataset. The most striking change was in precision, which increased from 0.53 to a perfect 1.00, indicating the elimination of false positives. This improvement is reflected in the confusion matrices, where the model trained on augmented data correctly classified all 56 non-defective samples, compared to 14 false positives in the baseline experiment.

Overall accuracy increased from 0.78 to 0.93, and the F1 score improved dramatically from 0.64 to 0.91. The recall saw a modest improvement from 0.80 to 0.84, indicating a slight reduction in false negatives. The ROC AUC also showed significant improvement, increasing from 0.899 to 0.973, suggesting better overall discrimination between classes.

\subsection{EfficientNetB0 Results}

\begin{table}[H]
\centering
\caption*{\textbf{Table 2}\\EfficientNetB0 Performance Comparison}
\vspace{-10pt}
\begin{tabular}{lcc}
\toprule
\textbf{Metric} & \textbf{RealData} & \textbf{AugmentedData} \\
\midrule
Accuracy & 0.8395 & 0.8485 \\
Precision & 1.00 & 1.00 \\
Recall & 0.35 & 0.65 \\
F1 Score & 0.52 & 0.79 \\
ROC AUC & 0.9672 & 0.9801 \\
\bottomrule
\end{tabular}
\end{table}

\begin{figure}[H]
\centering
\includegraphics[width=0.9\linewidth]{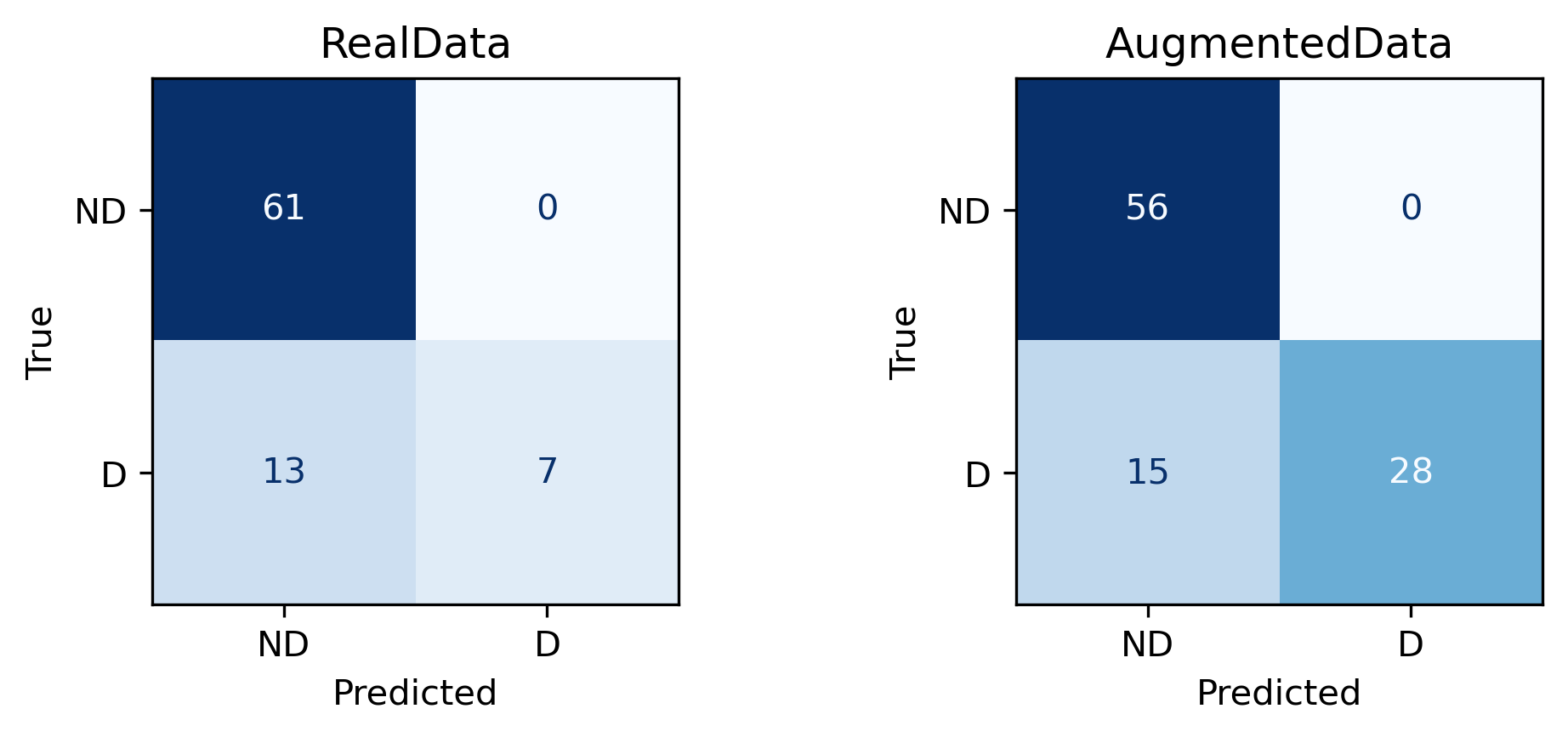}
\caption*{\textbf{Fig. 6.} EfficientNetB0 Confusion Matrices for RealData (left) and AugmentedData (right). ND = Non-Defective, D = Defective}
\end{figure}

\hspace{0.63cm}For the EfficientNetB0 model, the most significant improvement was in recall, which nearly doubled from 0.35 to 0.65. This indicates a substantial reduction in false negatives, with the model correctly identifying 65\% of defective samples when trained on the augmented dataset compared to only 35\% on the baseline dataset.

The precision remained perfect at 1.00 for both experiments, indicating no false positives in either case. The F1 score improved dramatically from 0.52 to 0.79, reflecting the improved balance between precision and recall. The ROC AUC also showed an improvement from 0.9672 to 0.9801, suggesting better discriminative capacity.

Overall accuracy showed a slight improvement from 0.8395 to 0.8485, which is notable given that the test set for the augmented experiment included a higher proportion of the more challenging defective class.

\subsection{MobileNetV2 Results}

\begin{table}[H]
\centering
\caption*{\textbf{Table 3}\\MobileNetV2 Performance Comparison}
\vspace{-10pt}
\begin{tabular}{lcc}
\toprule
\textbf{Metric} & \textbf{RealData} & \textbf{AugmentedData} \\
\midrule
Accuracy & 0.9012 & 0.8586 \\
Precision & 1.00 & 1.00 \\
Recall & 0.60 & 0.6744 \\
F1 Score & 0.75 & 0.8056 \\
ROC AUC & 0.9508 & 0.9805 \\
\bottomrule
\end{tabular}
\end{table}

\begin{figure}[H]
\centering
\includegraphics[width=0.9\linewidth]{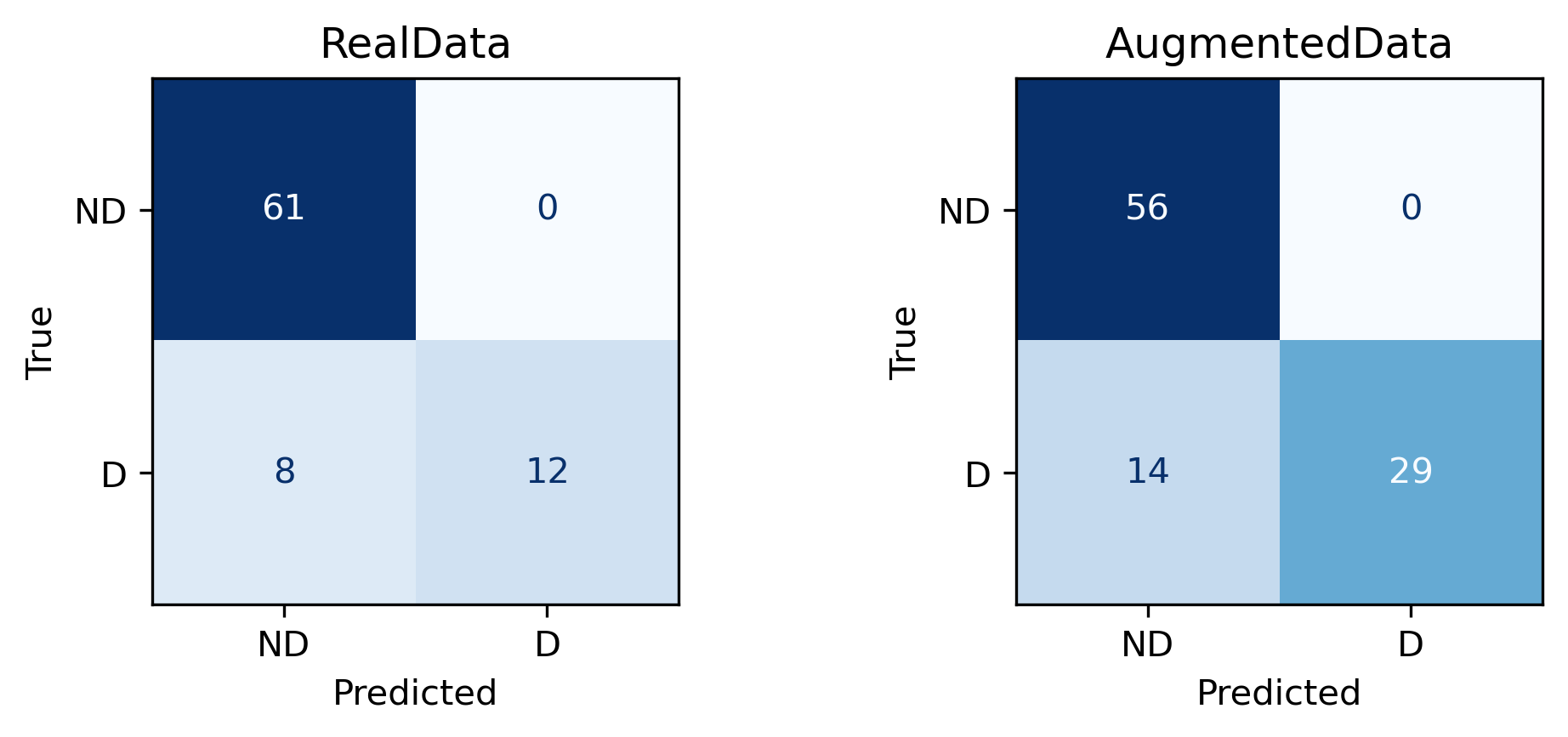}
\caption*{\textbf{Fig. 7.} MobileNetV2 Confusion Matrices for RealData (left) and AugmentedData (right). ND = Non-Defective, D = Defective}
\end{figure}

\hspace{0.63cm}The MobileNetV2 model showed improvement in defect detection capabilities when trained on the augmented dataset, particularly in terms of recall, which increased from 0.60 to 0.6744. This represents a reduction in false negatives, with more defective samples being correctly identified.

Similar to the other models, precision remained perfect at 1.00 for both experiments, indicating no false positives. The F1 score improved from 0.75 to 0.8056, reflecting the better balance between precision and recall. The ROC AUC showed significant improvement from 0.9508 to 0.9805, indicating enhanced discriminative ability.

Interestingly, the overall accuracy appeared to decrease slightly from 0.9012 to 0.8586. However, this metric should be interpreted with caution given the different class distributions in the validation sets and the higher priority placed on defect detection (minority class) in manufacturing contexts.

\subsection{Comparative Analysis Across Models}

\begin{figure}[H]
\centering
\includegraphics[width=0.9\linewidth]{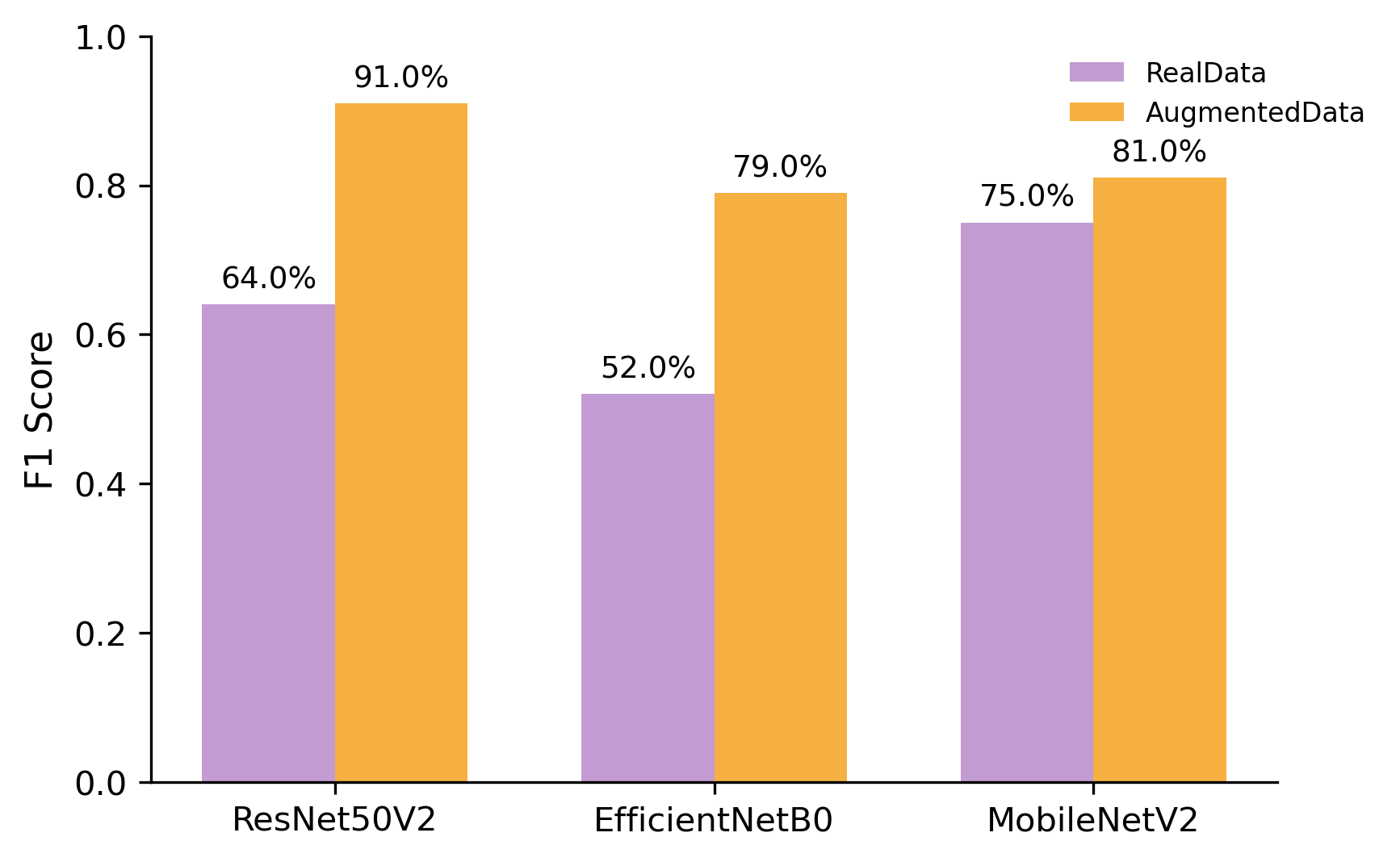}
\caption*{\textbf{Fig. 8.} F1 Score Comparison Across Models}
\end{figure}

\begin{figure}[H]
\centering
\includegraphics[width=0.9\linewidth]{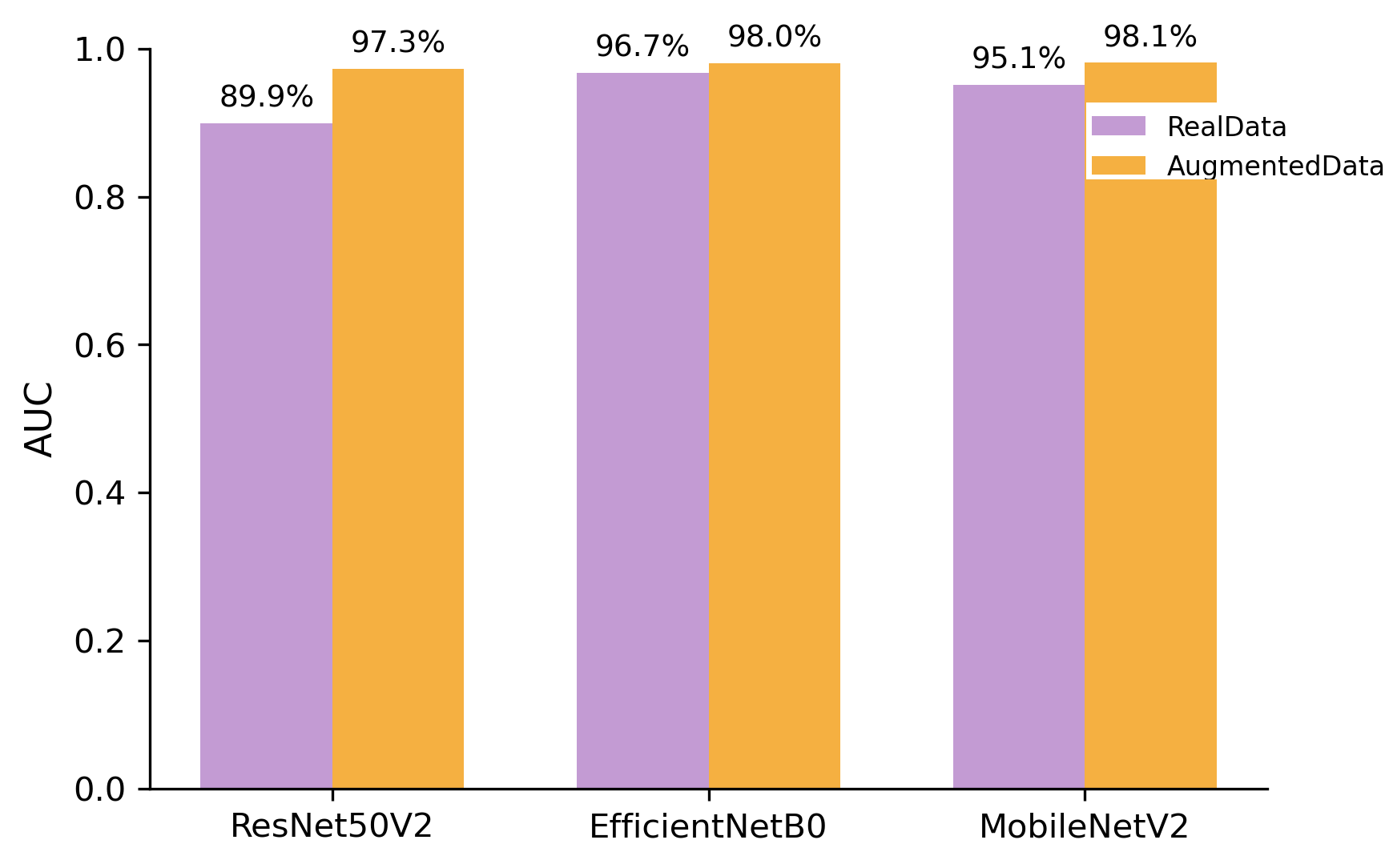}
\caption*{\textbf{Fig. 9.} ROC AUC Comparison Across Models}
\end{figure}

\hspace{0.63cm}All three models showed improvements when trained on the augmented dataset, particularly in metrics relevant to defect detection. The improvements in F1 score were most significant for ResNet50V2 and EfficientNetB0, with both showing a 0.27 point increase. MobileNetV2 showed a more modest improvement of 0.06 points, though it started from a higher baseline.

In terms of ROC AUC, all models showed improvements, with ResNet50V2 exhibiting the largest gain (+0.074), followed by MobileNetV2 (+0.030) and EfficientNetB0 (+0.013).

The most consistent improvement across all models was in recall, indicating enhanced ability to detect actual defects. This aligns with findings in other industrial domains where machine learning models have shown similar patterns of improvement when addressing class imbalance [28]. This is particularly important in manufacturing quality control, where missing a defect can have serious consequences.

Notably, all models maintained perfect precision (1.00) on the validation set when trained on the augmented dataset, meaning no false positives were generated. This is crucial in manufacturing contexts where false alarms can lead to unnecessary production stoppages and associated costs.

\section{Discussion}

\subsection{Impact of Synthetic Data Augmentation}

\hspace{0.63cm}The results clearly demonstrate that augmenting the training dataset with DDPM-generated synthetic images of defective glass bottles significantly improves model performance, particularly in detecting the minority class (defective products). This improvement was observed consistently across all three CNN architectures tested, though with varying degrees of enhancement.

Based on the experiments, several factors can be identified that likely contribute to this performance improvement: Improved class balance: By adding synthetic defective images, the imbalance ratio was reduced from approximately 3:1 to 1.7:1, which provided more balanced training signals to the models. Enhanced feature learning: The synthetic images introduced additional variations of defect patterns, helping the models learn more robust features for defect detection. Reduced overfitting: The larger and more diverse training set helped prevent the models from overfitting to the limited number of real defective samples. Preservation of defect characteristics: The DDPM demonstrated an ability to capture subtle patterns in the defective class, generating synthetic samples that maintained critical visual cues needed for accurate classification.

What appears most noteworthy is the consistent improvement in recall across all models, as this indicates better detection of actual defects—a critical capability in manufacturing quality control where missing defects can lead to product failures, customer complaints, and brand damage.

\subsubsection{Feature Space Analysis}

While performance metrics clearly demonstrate the benefits of using synthetic data, further investigation was conducted into how DDPM-generated images relate to real samples in terms of their feature representations. To achieve this, a feature space analysis using t-distributed Stochastic Neighbor Embedding (t-SNE) was performed. This technique helped visualize the high-dimensional feature representations in a two-dimensional space while preserving local relationships between data points.

For this analysis, feature vectors were extracted from all three image categories using a pre-trained ResNet50V2 model. These high-dimensional features were then projected into a two-dimensional space using t-SNE with perplexity = 30 and 2000 iterations (random\_state = 42) for stable, interpretable visualization, allowing for examination of the distribution patterns.

\begin{figure}[H]
\centering
\includegraphics[width=0.95\linewidth]{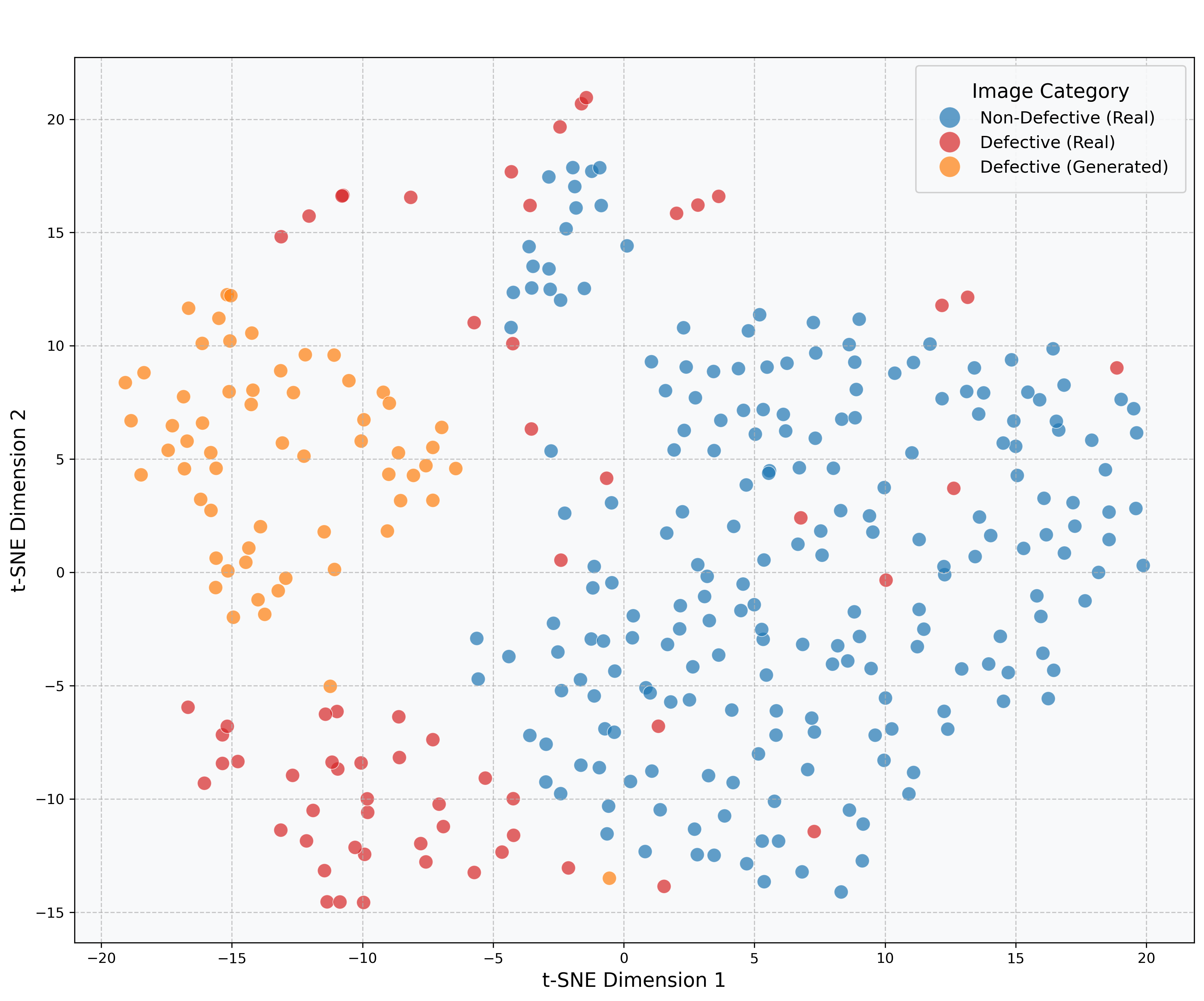}
\caption*{\textbf{Fig. 10.} t-SNE visualization of the dataset's feature vectors extracted using a pre-trained ResNet50V2 model. The plot reveals the distribution of real non-defective samples (blue), real defective samples (red), and DDPM-generated defective samples (orange) in the feature space}
\end{figure}

Upon examining Figure 10, several interesting patterns can be observed: Feature space clustering: The synthetic defective samples (orange) formed a distinct cluster separate from the non-defective samples (blue). This is particularly encouraging as it confirms that the synthetic images successfully captured discriminative features that differentiate defective from non-defective bottles. Complementary feature distribution: While the synthetic defective samples clustered separately from real defective samples (red), they occupied a region of the feature space that remained distinct from non-defective samples. This suggests that the synthetic samples weren't merely replicating existing defects but were capturing novel defect patterns that complemented the real data. Enhanced feature diversity: The distribution of synthetic samples within their cluster indicated that the DDPM had generated diverse variations of defects. This diversity likely contributed to the improved recall rates observed across all architectures.

This visual analysis reinforces the hypothesis that diffusion models can generate realistic defective glass bottle images while preserving the essential semantic features needed for accurate classification. The complementary nature of the synthetic samples in the feature space explains why their inclusion led to more robust models with improved detection performance.

\subsection{Advantages of Diffusion Models for Manufacturing Applications}

\hspace{0.63cm}The results highlight several advantages of using diffusion models for synthetic data generation in manufacturing quality control applications: High-fidelity generation: The DDPM-generated images maintained the visual characteristics of real defects, as evidenced by the improved classifier performance. This suggests that diffusion models can effectively capture the distinctive features of manufacturing defects. Training with limited data: Despite having only 63 real defective images for training the diffusion model, it successfully generated diverse and useful synthetic samples. This capacity to work with small datasets is particularly valuable in manufacturing settings where defective samples are scarce. Controlled augmentation: The approach enabled precise control over the level of data augmentation, allowing for reduction of the class imbalance to a specific target ratio (from 3:1 to 1.7:1) without introducing excessive synthetic data that might overwhelm the real samples. This efficient organization of synthetic data aligns with optimized data management strategies seen in other domains such as healthcare information systems [29]. Preservation of decision boundaries: The consistent maintenance of perfect precision across all models suggests that the synthetic images respected the natural decision boundaries between classes rather than introducing confusing patterns that might have led to false positives.

\subsection{Model-Specific Observations}

\hspace{0.63cm}Each model responded differently to the synthetic data augmentation:

\subsubsection{ResNet50V2}
ResNet50V2 showed the most dramatic improvement, with a 15 percentage point increase in accuracy (from 78\% to 93\%) and a substantial improvement in precision (from 0.53 to 1.00). This suggests that the deeper architecture of ResNet50V2 was better able to leverage the additional training data to refine its decision boundaries.

\subsubsection{EfficientNetB0}
EfficientNetB0 experienced the largest improvement in recall, nearly doubling from 35\% to 65\%. This indicates that the augmented dataset significantly enhanced the model's ability to identify defective samples, which it had previously struggled with despite having perfect precision.

\subsubsection{MobileNetV2}
MobileNetV2 showed more modest improvements overall but still benefited from the augmented dataset, particularly in terms of recall (60\% to 67.44\%) and F1 score (0.75 to 0.81). As the most lightweight model among the three, MobileNetV2's performance with the augmented dataset makes it an attractive option for resource-constrained deployment scenarios.

\subsection{Practical Implications for Manufacturing}

\hspace{0.63cm}The findings have several important implications for glass manufacturing and quality control: Cost-effective quality improvement: Synthetic data augmentation offers a cost-effective alternative to collecting large numbers of real defective samples, which would require extensive monitoring of production lines. Balanced precision and recall: The ability to maintain perfect precision while improving recall addresses the dual challenges in manufacturing quality control: minimizing false alarms while ensuring defect detection. Flexible model selection: The consistent improvement across architectures of varying complexity allows manufacturers to select models based on their specific computational constraints and performance requirements. Scalability: Once trained, the DDPM can generate additional synthetic samples as needed, allowing for further refinement of models or adaptation to new defect types.

\subsection{Limitations and Future Work}

\hspace{0.63cm}While the results are promising, several limitations and opportunities for future research should be acknowledged: Limited defect diversity: The current study focused on a binary classification (defective vs. non-defective) without distinguishing between different types of defects. Future work could explore multi-class classification to identify specific defect categories. Synthetic image quality: Although the DDPM-generated images were effective in improving model performance, further research could focus on enhancing the realism and diversity of synthetic defects through improved DDPM architectures or conditioning techniques. Real-world deployment: The study evaluated models on a held-out validation set rather than in a real production environment. Future work should assess the performance of these models in live manufacturing settings. Explainability: The current approach focuses on performance metrics without addressing the explainability of the models' decisions. Incorporating explainable AI techniques could provide insights into which features the models are using to identify defects. This would enhance trust in the system, similar to how prioritization and explanation techniques improve user feedback systems in other domains [31]. Advanced computational paradigms: While the current CNN architectures provided significant improvements, exploration of emerging computational approaches such as hyperdimensional computing could potentially offer additional benefits for processing high-dimensional image data in manufacturing environments, as demonstrated by Ghajari \textit{et al.} [30] in anomaly detection contexts. Integration with decision-making frameworks: Explore integrating diffusion-based augmentation with multi-criteria decision-making frameworks [32] to automatically rank and prioritize synthetic defect types for targeted model refinement. Generalization to other manufacturing domains: While focused on bottle glass manufacturing, the methodology could potentially be extended to other industries with similar class imbalance challenges. Future research could validate the approach in diverse manufacturing contexts.

\section{Conclusions}
\hspace{0.63cm}This study has demonstrated the effectiveness of using Denoising Diffusion Probabilistic Models (DDPMs) to generate synthetic defective glass bottle images for addressing class imbalance in manufacturing defect detection. By augmenting the training dataset with DDPM-generated images, significant improvements were achieved in the performance of three popular CNN architectures (ResNet50V2, EfficientNetB0, and MobileNetV2) in detecting defective products. The feature space analysis using t-SNE further validated that the synthetic images capture the essential characteristics that differentiate defective from non-defective samples, while introducing complementary variations that enhance model robustness.
Key improvements were observed in recall rates, which increased from 35\% to 65\% for EfficientNetB0, 60\% to 67\% for MobileNetV2, and 80\% to 84\% for ResNet50V2, all while maintaining perfect precision. The most dramatic overall improvement was observed in ResNet50V2's accuracy, which increased from 78\% to 93\% when trained with augmented data.
The results highlight the potential of diffusion models as a powerful tool for synthetic data generation in manufacturing quality control, offering a cost-effective approach to improving defect detection without requiring extensive collection of real defective samples. The methodology's effectiveness across different model architectures suggests its broad applicability and flexibility for different deployment scenarios.
Future work will extend this approach to multi-class defect classification, improve the quality and diversity of synthetic images, evaluate performance in real production environments, incorporate ex4plainability techniques, and validate the methodology in other manufacturing domains.
By addressing the critical challenge of class imbalance in defect detection, this research contributes meaningfully to the ongoing digital transformation of manufacturing, supporting the development of more efficient, accurate, and scalable quality control systems.

\thispagestyle{otherpagestyle}
\titleformat{\section}
{\bfseries\large}{\thesection}{}{}
\section*{Acknowledgement}

\vspace{-10pt}\hspace{-18pt}We acknowledge the use of the MVTec Anomaly Detection Dataset [23], which is made available under the Creative Commons Attribution-NonCommercial-ShareAlike 4.0 International License. This research was not funded by any grant.

\titleformat{\section}
{\bfseries\small}{\thesection}{}{}
\section*{Conflicts of Interest}

\vspace{-10pt}\hspace{-18pt}\small The authors declare no conflicts of interest.

\titleformat{\section}
{\bfseries\large}{\thesection}{1em}{}
\section*{References}

\vspace{-10pt}
\nocite{*}  
\printbibliography[heading=none]

@article{grandview2022, 

  author = {{Grand View Research}}, 

  title = {Glass Manufacturing Market to be Worth {USD} 167.50 Billion by 2030}, 

  journal = {PR Newswire}, 

  year = {2022}, 

  month = {December}, 

  day = {22}, 

  doi = {10.48550/prnewswire.2022.16750} 

}

@article{hütten2024, 

  author = {Hütten, N. and Gomes, M. A. and Hölken, F. and Andricevic, K. and Meyes, R. and Meisen, T.}, 

  title = {Deep Learning for Automated Visual Inspection in Manufacturing and Maintenance: A Survey of Open-Access Papers}, 

  journal = {Applied System Innovation}, 

  volume = {7}, 

  number = {1}, 

  pages = {11}, 

  year = {2024}, 

  month = {January}, 

  doi = {10.3390/asi7010011} 

}

@article{yang2020using, 

  author = {Yang, J. and Li, S. and Wang, Z. and Dong, H. and Wang, J. and Tang, S.}, 

  title = {Using Deep Learning to Detect Defects in Manufacturing: A Comprehensive Survey and Current Challenges}, 

  journal = {Materials}, 

  volume = {13}, 

  number = {24}, 

  pages = {5755}, 

  year = {2020}, 

  month = {December}, 

  doi = {10.3390/ma13245755} 

}

@article{wolleb2022diffusion, 

  author = {Wolleb, J. and Bieder, F. and Sandkühler, R. and Cattin, P. C.}, 

  title = {Diffusion Models for Medical Anomaly Detection}, 

  journal = {arXiv preprint arXiv:2203.04306v2}, 

  year = {2022}, 

  month = {March}, 

  doi = {10.48550/arXiv.2203.04306} 

}

@article{khazrak2024addressing, 

  author = {Khazrak, I. and Takhirova, S. and Rezaee, M. M. and Yadollahi, M. and Green II, R. C. and Niu, S.}, 

  title = {Addressing Small and Imbalanced Medical Image Datasets Using Generative Models: A Comparative Study of {DDPM} and {PGGANs} with Random and Greedy {K} Sampling}, 

  journal = {arXiv preprint}, 

  volume = {2412.12532}, 

  year = {2024}, 


  doi = {10.48550/arXiv.2412.12532} 

}

@online{weng2021diffusion, 

  author = {Weng, L.}, 

  title = {What are diffusion models?}, 

  year = {2021}, 

  month = {July}, 


  organization = {lilianweng.github.io}, 

  doi = {10.48550/arXiv.2110.02037} 

}

@article{groth2013glass, 

  author = {Groth, S. R. and Zhongfei, R.}, 

  title = {Glass Defect Detection Techniques using Digital Image Processing -- A Review}, 

  journal = {International Journal of Digital Applications \& Contemporary Research}, 

  volume = {1}, 

  number = {10}, 

  pages = {1--17}, 

  year = {2013}, 

  month = {May} 

  % DOI not available 

}

@article{malamas2003survey, 

  author = {Malamas, E. N. and Petrakis, E. G. M. and Zervakis, M. and Petit, L. and Legat, J.-D.}, 

  title = {A survey on industrial vision systems, applications and tools}, 

  journal = {Image and Vision Computing}, 

  volume = {21}, 

  number = {2}, 

  pages = {171--188}, 

  year = {2003}, 

  month = {February}, 

  doi = {10.1016/S0262-8856(02)00152-X} 

}

@article{zhou2020surface, 

  author = {Zhou, X. and Wang, Y. and Zhu, Q. and Mao, J. and Xiao, C. and Lu, X. and Zhang, H.}, 

  title = {A Surface Defect Detection Framework for Glass Bottle Bottom Using Visual Attention Model and Wavelet Transform}, 

  journal = {IEEE Transactions on Industrial Informatics}, 

  volume = {16}, 

  number = {4}, 

  pages = {2189--2201}, 

  year = {2020}, 

  doi = {10.1109/TII.2019.2935153} 

}

@article{buda2018systematic, 

  author = {Buda, M. and Maki, A. and Mazurowski, M. A.}, 

  title = {A systematic study of the class imbalance problem in convolutional neural networks}, 

  journal = {Neural Networks}, 

  volume = {106}, 

  pages = {249--259}, 

  year = {2018}, 

  doi = {10.1016/j.neunet.2018.07.011} 

}

@article{khan2018cost, 

  author = {Khan, S. H. and Hayat, M. and Bennamoun, M. and Sohel, F. A. and Togneri, R.}, 

  title = {Cost-sensitive learning of deep feature representations from imbalanced data}, 

  journal = {IEEE Transactions on Neural Networks and Learning Systems}, 

  volume = {29}, 

  number = {8}, 

  pages = {3573--3587}, 

  year = {2018}, 

  doi = {10.1109/TNNLS.2017.2732482} 

}

@article{liu2015ensemble, 

  author = {Liu, X. and Lin, S. and Fang, J. and Xu, Z.}, 

  title = {Is extreme learning machine feasible? A theoretical assessment (Part I)}, 

  journal = {IEEE Transactions on Neural Networks and Learning Systems}, 

  volume = {26}, 

  number = {1}, 

  pages = {7--20}, 

  year = {2015}, 

  month = {January}, 

  doi = {10.1109/TNNLS.2014.2308135} 

}

@article{shorten2019survey, 

  author = {Shorten, C. and Khoshgoftaar, T. M.}, 

  title = {A survey on image data augmentation for deep learning}, 

  journal = {Journal of Big Data}, 

  volume = {6}, 

  number = {1}, 

  pages = {60}, 

  year = {2019}, 

  doi = {10.1186/s40537-019-0197-0} 

}

@article{adami2025, 

  author = {Adami, B. and Karimian, N.}, 

  title = {{rPPG-SysDiaGAN}: Systolic-Diastolic Feature Localization in {rPPG} Using Generative Adversarial Network with Multi-Domain Discriminator}, 

  journal = {arXiv preprint arXiv:2504.01220}, 

  year = {2025}, 

  month = {April}, 

  doi = {10.48550/arXiv.2504.01220} 

}

@inproceedings{ho2020denoising, 

  author    = {Ho, Jonathan and Jain, Ajay and Abbeel, Pieter}, 

  title     = {Denoising Diffusion Probabilistic Models}, 

  booktitle = {Advances in Neural Information Processing Systems}, 

  volume    = {33}, 

  pages     = {6840--6851}, 

  year      = {2020}, 

  doi       = {10.48550/arXiv.2006.11239} 

}

@inproceedings{nichol2021improved, 

  author    = {Nichol, Alexander Quinn and Dhariwal, Prafulla}, 

  title     = {Improved Denoising Diffusion Probabilistic Models}, 

  booktitle = {Proceedings of the 38th International Conference on Machine Learning}, 

  pages     = {8162--8171}, 

  year      = {2021}, 

  doi       = {10.48550/arXiv.2102.09672} 

}

@inproceedings{dhariwal2021diffusion, 

  author    = {Dhariwal, Prafulla and Nichol, Alex}, 

  title     = {Diffusion Models Beat GANs on Image Synthesis}, 

  booktitle = {Advances in Neural Information Processing Systems}, 

  volume    = {34}, 

  pages     = {8780--8794}, 

  year      = {2021}, 

  doi       = {10.48550/arXiv.2105.05233} 

}

@inproceedings{karras2022, 

  author    = {Karras, Tero and Aittala, Miika and Aila, Timo and Laine, Samuli}, 

  title     = {Elucidating the Design Space of Diffusion-Based Generative Models}, 

  booktitle = {Advances in Neural Information Processing Systems}, 

  volume    = {35}, 

  pages     = {26565--26577}, 

  year      = {2022}, 

  doi       = {10.48550/arXiv.2206.00364} 

}

@inproceedings{rombach2022, 

  author    = {Rombach, Robin and Blattmann, Andreas and Lorenz, Dominik and Esser, Patrick and Ommer, Björn}, 

  title     = {High-Resolution Image Synthesis with Latent Diffusion Models}, 

  booktitle = {Proceedings of the IEEE/CVF Conference on Computer Vision and Pattern Recognition}, 

  pages     = {10684--10695}, 

  year      = {2022}, 

  doi       = {10.48550/arXiv.2112.10752} 

}

@article{weimer2016transfer, 

  author  = {Weimer, Daniel and Scholz-Reiter, Bernd and Shpitalni, Moshe}, 

  title   = {Design of Deep Convolutional Neural Network Architectures for Automated Feature Extraction in Industrial Inspection}, 

  journal = {CIRP Annals}, 

  volume  = {65}, 

  number  = {1}, 

  pages   = {417--420}, 

  year    = {2016}, 

  doi     = {10.1016/j.cirp.2016.04.072} 

}

@article{ren2021glass, 

  author  = {Ren, Kui and Zheng, Tianhang and Qin, Zhan and Liu, Xue}, 

  title   = {Adversarial Attacks and Defenses in Deep Learning}, 

  journal = {Engineering}, 

  volume  = {6}, 

  number  = {3}, 

  pages   = {346--360}, 

  year    = {2020}, 

  doi     = {10.1016/j.eng.2019.12.012} 

}

@inproceedings{sandler2018mobilenetv2, 

  author    = {Sandler, Mark and Howard, Andrew and Zhu, Menglong and Zhmoginov, Andrey and Chen, Liang-Chieh}, 

  title     = {MobileNetV2: Inverted Residuals and Linear Bottlenecks}, 

  booktitle = {Proceedings of the IEEE Conference on Computer Vision and Pattern Recognition}, 

  pages     = {4510--4520}, 

  year      = {2018}, 

  doi       = {10.1109/CVPR.2018.00474} 

}

@inproceedings{tan2019efficientnet, 

  author    = {Tan, Mingxing and Le, Quoc V.}, 

  title     = {EfficientNet: Rethinking Model Scaling for Convolutional Neural Networks}, 

  booktitle = {Proceedings of the 36th International Conference on Machine Learning}, 

  pages     = {6105--6114}, 

  year      = {2019}, 

  doi       = {10.48550/arXiv.1905.11946} 

}

@article{ferguson2018detection, 

  author  = {Ferguson, Matthew and Ak, Ramin and Lee, Yung-Tsun Tina and Law, Kincho H.}, 

  title   = {Detection and Segmentation of Manufacturing Defects with Convolutional Neural Networks and Transfer Learning}, 

  journal = {Smart and Sustainable Manufacturing Systems}, 

  volume  = {2}, 

  number  = {1}, 

  pages   = {137--164}, 

  year    = {2018}, 

  doi     = {10.1520/SSMS20170036} 

}

@inproceedings{he2016identity, 

  author    = {He, Kaiming and Zhang, Xiangyu and Ren, Shaoqing and Sun, Jian}, 

  title     = {Identity Mappings in Deep Residual Networks}, 

  booktitle = {European Conference on Computer Vision}, 

  pages     = {630--645}, 

  year      = {2016}, 

  doi       = {10.1007/978-3-319-46493-0_38} 

}

@inproceedings{bergmann2019mvtec, 

  author    = {Bergmann, Paul and Fauser, Michael and Sattlegger, David and Steger, Carsten}, 

  title     = {MVTec AD—A Comprehensive Real-World Dataset for Unsupervised Anomaly Detection}, 

  booktitle = {Proceedings of the IEEE/CVF Conference on Computer Vision and Pattern Recognition}, 

  pages     = {9592--9600}, 

  year      = {2019}, 

  doi       = {10.1109/CVPR.2019.00982} 

}

@online{zare2023, 

  author = {Zare, S. and Sun, Y.}, 

  title  = {Understanding Human Motion Intention from Motor Imagery EEG Based on Convolutional Neural Network}, 

  year   = {2023}, 

  url    = {https://ssrn.com/abstract=5005300}, 

  note   = {Available at SSRN 5005300} 

}

@article{mohammadagha2025, 

  author  = {Mohammadagha, M. and Najafi, M. and Kaushal, V. and Jibreen, A. M. A.}, 

  title   = {Machine Learning Models for Reinforced Concrete Pipes Condition Prediction: The State-of-the-Art Using Artificial Neural Networks and Multiple Linear Regression in a Wisconsin Case Study}, 

  journal = {arXiv preprint arXiv:2502.00363}, 

  year    = {2025}, 

  month   = {February}, 


  doi     = {10.48550/arXiv.2502.00363} 

}

@article{soltanmohammadi2025, 

  author  = {Soltanmohammadi, E. and Hikmet, N. and Akgun, D.}, 

  title   = {Tailored Partitioning for Healthcare Big Data: A Novel Technique for Efficient Data Management and Hash Retrieval in RDBMS Relational Architectures}, 

  journal = {Journal of Data Analysis and Information Processing}, 

  volume  = {13}, 

  number  = {1}, 

  pages   = {21--35}, 

  year    = {2025}, 

  month   = {February}, 


  doi     = {10.4236/jdaip.2025.131003} 

}

@article{ghajari2025, 

  author  = {Ghajari, G. and Ghimire, A. and Ghajari, E. and Amsaad, F.}, 

  title   = {Network Anomaly Detection for IoT Using Hyperdimensional Computing on NSL-KDD}, 

  journal = {arXiv preprint arXiv:2503.03031}, 

  year    = {2025}, 

  month   = {March}, 

  doi     = {10.48550/arXiv.2503.03031} 

}

@article{jafari2025, 

  author    = {Jafari, M. and Majidi, F. and Heydarnoori, A.}, 

  title     = {Prioritizing App Reviews for Developer Responses on Google Play}, 

  journal = {arXiv preprint arXiv:2502.01520}, 

  year      = {2025},

  doi       = {10.48550/arXiv.2502.01520} 

}

@online{deldadehasl2025, 

  author       = {Deldadehasl, M. and Hajian Karahroodi, H. and Haddadian Nekah, P.}, 

  title        = {Customer Clustering and Marketing Optimization in Hospitality: A Hybrid Data Mining and Decision-Making Approach from an Emerging Economy}, 

  year         = {2025}, 

  month        = {April}, 

  organization = {Preprints.org},

  doi       = {10.20944/preprints202504.0297.v1} 

  % DOI not available 

}

\end{document}